\documentclass[preprint,12pt]{elsarticle}

\usepackage{amssymb}
\usepackage{amsmath}
\usepackage{algorithmic}
\usepackage{graphicx}
\usepackage{textcomp}
\usepackage{xcolor}
\usepackage[ruled,vlined]{algorithm2e}
\def\BibTeX{{\rm B\kern-.05em{\sc i\kern-.025em b}\kern-.08em
    T\kern-.1667em\lower.7ex\hbox{E}\kern-.125emX}}
\usepackage{caption}
    
\usepackage{tabularx}
\usepackage[figuresright]{rotating}

\begin{document}

\begin{frontmatter}



\title{A Novel Non-population-based Meta-heuristic Optimizer Inspired by the Philosophy of Yi Jing}


\author[inst1]{Ho-Kin Tang}

\affiliation[inst1]{organization={Shenzhen JL Computational Science and Applied Research Institute},
            addressline={Level 6, Block 26, HongShan 6979 Phase Two, Longhua New District}, 
            city={Shenzhen},
            postcode={518131}, 
            state={GuangDong Province},
            country={China}}
\ead{hokintang@csar.ac.cn}

\author[inst2]{Sim Kuan Goh}

\affiliation[inst2]{organization={Air Traffic Management Research Institute, Nanyang Technological University},
            addressline={65 Nanyang Drive}, 
            postcode={637460}, 
            country={Singapore}}
\ead{skgoh@ntu.edu.sg}

\begin{abstract}
Drawing inspiration from the philosophy of Yi Jing, Yin-Yang pair optimization~(YYPO) has been shown to achieve competitive performance in single objective optimizations. Besides, it has the advantage of low time complexity when comparing to other population-based optimization. As a conceptual extension of YYPO, we proposed the novel Yi optimization~(YI) algorithm as one of the best non-population-based optimizer. Incorporating both the harmony and reversal concept of Yi Jing, we replace the Yin-Yang pair with a Yi-point, in which we utilize the $\mathrm{L\acute{e}vy}$ flight to update the solution and balance both the effort of the exploration and the exploitation in the optimization process. As a conceptual prototype, we examine YI with IEEE CEC 2017 benchmark and compare its performance with a $\mathrm{L\acute{e}vy}$ flight-based optimizer CV1.0, the state-of-the-art dynamical Yin-Yang pair optimization in YYPO family and a few classical optimizers. According to the experimental results, YI shows highly competitive performance while keeping the low time complexity. Hence, the results of this work have implications for enhancing meta-heuristic optimizer using the philosophy of Yi Jing, which deserves research attention.
\end{abstract}



\begin{keyword}
Yi optimization \sep Yin-Yang pair optimization \sep $\mathrm{L\acute{e}vy}$ flight \sep CEC2017 \sep numerical optimization
\end{keyword}

\end{frontmatter}


\section{Introduction}
In the long contest of evolution pressure in nature, living beings have developed different survival strategies. The retaining strategies being adapted are the ones that pass on generations through natural selection. They are often highly optimized as a result of million years of evolution. These brilliant strategies have been a great source of inspiration for a variety of heuristic optimization techniques~\cite{mirjalili2020genetic}, for examples, the cuckoo breeding behavior inspired the cuckoo search~\cite{Yang2009}, the hunting strategy of a grey wolf pack inspired the Grey Wolf search~\cite{mirjalili2014grey}, the evolution principle of DNA inspired the Genetic Adaptive approaches\cite{brindle1980genetic}. Among living species, human is unique, reflected by the unprecedented accumulation in cultural and philosophical contexts. These immortal gems represent the trials of humans to interpret nature. The retaining philosophies, similar to survival strategies, also passed through the contest of time. Some of the ideas are sophisticated and inspired the development of meta-heuristic optimization techniques and related applications~\cite{al2019survey,sinha2017review,fernandez2019evolutionary}. For example, the ancient Yin-Yang idea inspired the non-population-based Yin-Yang pair optimization~(YYPO)~\cite{Punnathanam2016,Punnathanam2016a,Maharana2017,Punnathanam2019,Punnathanam2019a}. The technique is advantageous of low time complexity and being competitive in optimization problems. It has been successfully implemented in some real-time optimization problems, in which the low time-complexity of the algorithm is important~\cite{Punnathanam2017,Heidari2017,Yang2018,Song2020}. 

\begin{figure*}
    \centering
    \includegraphics[width=0.9\textwidth]{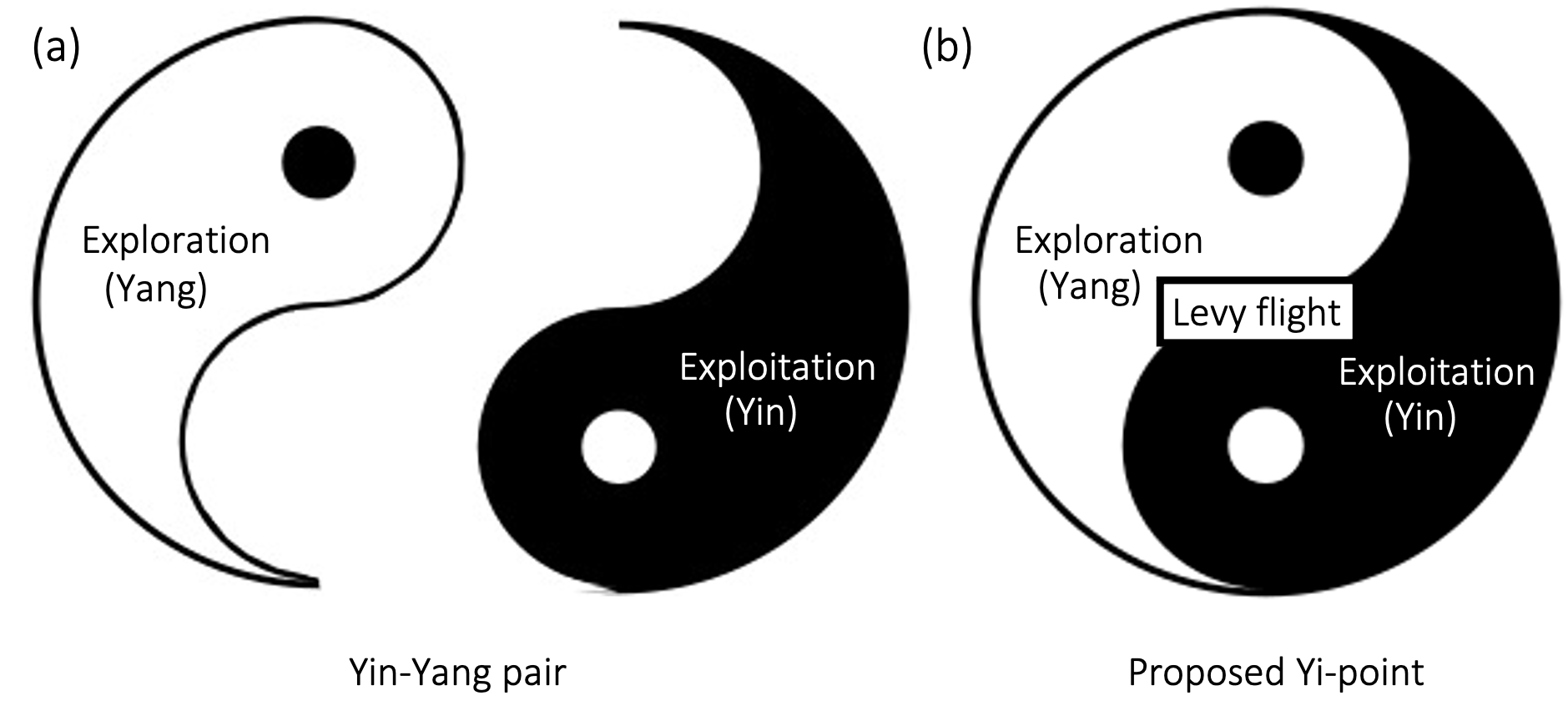}
    \caption{Illustration of the conceptual difference between the Yin-Yang pair optimization~(YYPO) and the Yi algorithm~(YI). $\mathrm{L\acute{e}vy}$ flight is a technique that combines both functions of exploration and exploitation in the yin-yang pair optimization.}
    \label{fig:fig1}
\end{figure*}

Yi Jing, also known as The Book of Changes, carries the profound philosophy of harmony and unity. The Yi Jing script reads "Taiji generates two complementary forces, two complementary forces generate four aggregates...". In ancient time, the ancestors try to use the binary hierarchy in describing the phenomenon in our universe. Inspired by the complementary forces of Yin and Yang, YYPO uses the idea of harmony to design the Yin-Yang pair balancing the exploration and the exploitation in the optimization. In the pair, one point favors exploration~(Yan), while another point favors exploitation~(Ying). Two points cooperate in the search space to find the optimized solution.

In this paper, we are inspired by an important aspect of the philosophy of Yi Jing -- the reversal property: Taiji can generate Yin and Yang, and Yin and Yang can also combine back to form Taiji. These are the infinite cycles of changes. Inspired by the reversal property between Yin-Yang and Taiji, we unify the Yin-Yang pair into a single Yi-point to achieve an enhanced balance between the exploration process and the exploitation process. We proposed a highly competitive non-population-based Yi algorithm~(YI) based on the YYPO framework, but replacing the Yin-Yang pair with the Yi-point empowered by the $\mathrm{L\acute{e}vy}$ flight.

As YI is a newly proposed algorithm, we like to compare it with a few classical nature-inspired algorithms. When we compare YI with other algorithms, we find YI shares some ingredients with others, but also have its unique features. We first compare with YYPO and its improved version of dynamical YYPO. As its variant, we inherit its point-based feature and its dynamical archive. The main difference is that we reduce the number of points used in the searching from two to one, replacing the Yin-Yang pair with the Yi-point, this further enhances the low time complexity as well as improving the control over the exploration and exploitation. Also, instead of using both one-way splitting and D-way splitting of updating in YYPO, YI use the unified $\mathrm{L\acute{e}vy}$ flight updating. 

When comparing with the famous population-based methods like particle swarm optimization (PSO)~\cite{kennedy1995particle,shi1998modified} and genetic algorithm~(GA), the YI and YYPO family share some insights from these algorithms. Similar to PSO memory of its global best point~(gbest), we have temporary memory of our YI point or the Ying-yang pair during the archive time. However, different from PSO, the memory does not affect the decision of searching, as the search in YI is using the Levy flight process that does not have the velocity affected by the memory. Moreover, YI also shares the elitism concept with GA, in which we continue the search with the point of best fitness from the record every successive achieve interval. But YI does not carry out any breeding between points in the population. 

To scrutinize the proposed YI, we use the benchmark from CEC2017, which contains 29 challenging functions for optimization~\cite{wu2017problem}. We compare YI against the dynamical YYPO~(dYYPO)\cite{Maharana2017}, CV1.0 and a few classical algorithms (i.e, diffential evolution, PSO, simulated annealing). We find that YI  outperforms the dYYPO, CV1.0 and the classical algorithms in most of the functions. This paper is organized as the following. In section II, we describe the proposed Yi algorithm. In section III, we outline the setting of the CEC2017 experiment and its result on 10D, 30D, and 50D. In section IV, we discuss and compare the algorithm performance between the result from the Yi algorithm, the dynamical YYPO~(dYYPO), the Cuckoo search~(CV1.0) and the classical algorithms, and give conclusions.

\section{Algorithm Descriptions}
In this section, we discuss the algorithm of the proposed YI. As a variant of YYPO, we first review the concept of the Yin-Yang pair optimization. Then we discuss two main elements of YI. (1) the archive framework that we adapt and improve from dYYPO, (2) the newly proposed Yi-point, that perfectly performs the function of both exploration~(Yang) and exploitation~(Yin) in the Yin-Yang pair. Finally, we outline the detail of YI, and give the pseudocode. 

\subsection{Review of the Yin-Yang pair optimization}
The Yin-Yang pair optimization is inspired by the duality of opposite force conflicting in nature depicted in the Yi Jing. One aspect gradually changes to the other and the cycle of changing continues forever. During the change, if one could foster the balance between two forces, it results in harmony. In the field of evolutionary computing, two conflicting behaviors are exploitation and exploration. Inspired by the philosophy, Varun et al. proposed the YYPO that makes use of the concept illustrated in Fig.~\ref{fig:fig1}(a)\cite{Punnathanam2016}. 

YYPO algorithm uses two points~(P1 and P2) to do the optimization task. P1 is designed to focus on exploitation, and P2 is designed to focus on exploration. P1 and P2 act as centers to explore the variable space, with searching radii $\delta_1$ and $\delta_2$, respectively. Initially, both $\delta_1$ and $\delta_2$ are set to be 0.5. During the updating process, the searching radii $\delta_1$ of P1 decreases to facilitate the exploitation task, while $\delta_2$ of P2 increases to enhance the exploration ability. YYPO mainly consists of two stages, splitting stage and archive stage. In the splitting stage, they use two strategies that being selected randomly, namely one-way splitting and D-way splitting. One-way splitting is directional splitting using the vector with all positive element or all negative element, while D-way splitting use the vector with elements of random sign. The probabilities of choosing a splitting strategy have been optimized~\cite{Punnathanam2016}.

YI does not inherit any feature of YYPO in the splitting stage and its Yin-yang pair, so we rename our algorithm by YI. But we share partly the structure of the archive stage, in which we will discuss in the next section. 

\subsection{Dynamical archive}
In the successive development of YYPO, the dynamical YYPO used dynamical approach to achieve better control over the exploration and the exploitation during different intervals in the search. YI uses the framework of dynamical archive, but a different strategy of updating when comparing to dYYPO. We divide the whole optimization process into $n_{T} = I_{max}-I_{min}+1$ intervals according to the number of function evaluations done. Starting from $I_{max}$, $I$ descended by one every time when entering the next interval, until reaching $I_{min}$ at the end of the search. At the start of the search, the search process favors the strategy of exploration, so a large $I$ is preferred, allowing the Yi-point to explore the function space without the hindrance of starting from the best obtained solution. In the ending phase of searching, the exploitation becomes more important, so a smaller $I$ allows the Yi-point to focus on searching for the best solution locally. On the contrary, dYYPO uses ascending $I$. It might be due to the fact that a smaller $I$ in the phase of exploration improves the communication between the pair, which avoids wasting too much time on exploiting near bad solutions.

\subsection{The inspired Yi-point}
The Yi Jing promotes the concept of the infinite cycles of changes in all observing phenomenon. This requires the reversal property of the Yin-Yang.: Taiji can generate Yin and Yang, and Yin and Yang can also combine back to form Taiji. Inspired by the reversal property between Yin-Yang and Taiji, we combine the two points in Yin-yang pair back to a unified point, we call it Yi-point, which represents Taiji. The main concept and the relation are illustrated in Fig.~\ref{fig:fig1}. The Yi-point plays a central role in the algorithm. The strategy of updating the Yi-point needs to strike a balance between exploration and exploitation. In our scheme, we use the profound $\mathrm{L\acute{e}vy}$ flight to perform the task. The $\mathrm{L\acute{e}vy}$ flight has been found to capture the characteristic of the flight trajectories of many animals and insects~\cite{yao1999evolutionary, brown2007levy,pavlyukevich2007levy}, in balancing the effort of exploration and exploitation, to maximize the efficiency of searching foods. The following equations  describe the flight.

\begin{align}
    p_{new} &= p_0 + \alpha \otimes \delta_{\mathrm L\acute{e}vy}(\lambda)
\end{align}

\noindent where $\delta_{\mathrm L\acute{e}vy}$ are the random number generated from $\mathrm L\acute{e}vy$ distribution with stability parameter $\lambda$, $p_0$ and $p_{new}$ are the solutions before and after the splitting. The $\mathrm{L\acute{e}vy}$ distribution is fat-tail function. We choose $\lambda=1.5$ for the distribution, which provides the chance of far-field exploration without losing intensified local exploitation. In YI, we control the scope of the flight by $\alpha$, initialized as the dimension $D$. We emphasize that our proposed $\mathrm{L\acute{e}vy}$ flight is just one of the strategies to update the Yi-point, other schemes are not yet explored.

\subsection{Outline of the YI algorithm}
YI combines both dynamical archive and inspired Yi-point into a highly competitive optimizer with low time complexity. YI has three user-defined parameters, $I_{min}$, $I_{max}$ and $\sigma$. $I_{min}$ and $I_{max}$ are the minimum and the maximum value of $I$ that defines the number of splitting performed before picking the best from the archive. $\sigma$ is the decay rate of the scope of $\mathrm{L\acute{e}vy}$ flight.

As shown in the pseudo-code, we first initialize the Yi-point randomly in the bounded space and calculate its fitness. We initialize the archive duration $I=I_{max}$, and the search scope $\alpha=D$. We define the next interval $T_t$, once the iteration time $T$ exceeds $T_t$, we rescale the scope by dividing it by $\sigma$, and decrease the archive duration $I$. As the search proceeds, we perform more local searches and with shorter archive duration. During each archive duration, we carry out $I$ time of splitting and archive process. In each splitting process, we do the $\mathrm{L\acute{e}vy}$ splitting. If the updated element of the vector is out of the boundary, we choose a random position for this element. We record the best $p_{best}$ among the splitted $p_{new}$. Even the fitness of $p_{best}$ is worse than the fitness of $p_0$, we would still archive the $p_{best}$. To maintain the elitism of the Yi-point, we will choose the best to be $p_0$ in the archive at the end of each archive duration. We then empty the archive and start and new archive duration. The search repeats until the stopping criteria are met. At the end of the search, archive duration $I$ drops to $I_{min}$ after $I_{max}-I_{min}$ time of rescaling $\alpha$. We provide the pseudocode of the Yi algorithm in the text, including the main algorithm and the splitting function.

\begin{algorithm}
\SetAlgoLined
 Initialize the Yi-point $p$ randomly;\\
 Calculate the fitness of $p$, $f(p)$;\\
 $I = I_{max}$;\\
 $\alpha=D$, the dimension;\\
 Set $j=1, i=0$;\\
 Assign the next-interval time  $T_t= T_{max}/(I_{max}-I_{min}+1)$;\\
 \While{stopping criteria}{
 \If{$T>T_t$}
 {$\alpha= \alpha/\sigma$;\\
  $I = I_{max}-j$;\\
 $j=j+1$;\\
 $T_t= T_{max}\times j/(I_{max}-I_{min}+1)$;\\
 }
  $\mathrm{L\acute{e}vy}$ splitting of $p$ give $p_{best}$;\\
  Archive $p_{best}$ in $\{p_{rec}\}$;\\
  Set $p = p_{best};$\\
  i = i+1;\\
  \If{i equal to I}{
    $p = min(p_{rec})$;\\
    Empty the archive;\\
    Set $i=0$;\\
   }
 }
 Output $p = min(p_{rec})$ 
 \caption{Yi algorithm}
\end{algorithm}

\begin{algorithm}
\SetAlgoLined
 Input $p_0$;\\
 \For{$k=1$ to $2\times D$}{
  $\mathrm{L\acute{e}vy}$ flight process $p_k = p_0 + \alpha \otimes \delta_{Cauchy}$;
 }
 Choose the best of $p_k$ to output
 \caption{$\mathrm{L\acute{e}vy}$ splitting function}
\end{algorithm}

\section{Experiment and Results}
This section first describes the benchmark data-set from CEC 2017 that is used to assess our proposed method compared against dYYPO and CV1.0. The main properties of these algorithms are shown in Table~\ref{tab:compare}. Next, we discuss the parameters of the proposed method. Subsequently, we describe the experiment results and show the statistical tests. We also provide the time complexity analysis.

\subsection{Benchmark}
To validate our algorithm, we use the CEC 2017 benchmark: Single Objective Bound Constrained Real-Parameter Numerical Optimization~\cite{wu2017problem}, which comprises four main groups of functions, namely, unimodal functions~(F1,F3), simple multimodal functions~(F4-F10), hybrid functions~(F11-F20) and composition functions~(F21-F30). In total, there are 29 functions with diverse search spaces.

\subsection{Parameters}
In all the searches performed in this paper, the parameters are chosen in consideration of convergence in limited times of function evaluation. We use $I_{min}=6$, $I_{max}=15$ to divide the whole process into 11 intervals. The number of intervals cannot be too small, as this ensures the desired scope of the $\mathrm{L\acute{e}vy}$ flight at the end of the search to perform good exploitation tasks. For the decay rate, we choose $\sigma=3$. The search scope increases with the dimensions of the problem to secure the exploration coverage in the function space.

\subsection{Experimental Settings}
As a prototype of the new YI concept, we compare our method with the seminal version of the differential evoluation~(DE), the particle swarm optimization~(PSO), and the simulated annealing~(SA), as well as a $\mathrm{L\acute{e}vy}$ flight-based optimizer CV1.0, and the state-of-the-art dynamical YYPO~(dYYPO) in YYPO family. We followed their experimental settings. Both papers used the stopping criteria of $10000\times D$ function evaluations and each benchmark problem was run 51 times. We used the same numbers for our experiment to facilitate fair comparison. 

\subsection{Statistical Results}
 We also presented our experiment results in terms of best, worst, mean and standard deviation of the error values. The error values are computed by calculating the difference expected and the desired solution. Table~\ref{tab:10D},~\ref{table:30D} and~\ref{table:50D} showed the statistics of our experimental results for the test problems, for dimension $D = 10,30,50$, respectively.

\subsection{Algorithms}
To quantitatively assess our methods, we compare our method against DE, PSO, SA, CV1.0 and dYYPO algorithms, to examine the competitiveness of our proposed method. We compare the main features of these algorithms in Table~\ref{tab:compare}, and briefly discuss the algorithms in the following texts, for details please refer to the reference.

DE is a simple and one of the most powerful population-based heuristic for global optimization~\cite{storn1997differential}. It is a popular variant of evolutionary algorithms (EA) for multi-dimensional real-valued search spaces. Algorithmically, DE modified the mutation strategies of EA. DE generates new vectors (i.e., solutions) by summing the weighted difference between two population vectors to a third vector. Despite its simplicity, it demonstrates robustness to multi-modal problems and complex constrained optimization problems~\cite{10.3389/fbuil.2020.00102}.

PSO is a population based metaheuristic algorithm inspired by the behavior of how a flock of birds move together~\cite{kennedy1995particle,shi1998modified}. The movement of each bird, like a particle, is affected by their own experience and other birds experience. For each bird, they have their own experience best fitness locally (lbest). And the global best fitness~(gbest) is defined as the best point found among all birds. The position of gbest and lbest affect the movement of the birds in the update process, in term of the social adjustment by gbest, the self adjustment by lbest and the inertia.

SA is a stochastic global search optimization algorithm inspired by the thermodynamic property in materials~\cite{kirkpatrick1983optimization}. When the temperature is high, the system can be excited to different states without constraints of energy barriers. While at low temperatures, the system could be trapped in the state with a local minimum in energy. SA proposes the new solution and accepts it according to the Boltzmann weight of fitness by using the metropolis algorithm. So, in the beginning of the search at a high temperature, we can explore variable phase space freely. The temperature then decreases slowly to force the search to converge to a minimum.  

A number of studies have used $\mathrm{L\acute{e}vy}$ flight process to enhance the performance of the optimizer~\cite{jensi2016enhanced,chegini2018psoscalf,abdulwahab2019enhanced,amirsadri2018levy}. CV1.0~\cite{salgotra2018new} was one of them as a variant of the cuckoo search~\cite{pavlyukevich2007levy}. The Cauchy-based global search was used in the initial phase for exploration, then the Grey wolf optimization algorithm~(GWO)~\cite{mirjalili2014grey} was utilized in the remaining phase for exploitation.

We have chosen the seminal implementation of DE, PSO and SA that are provided in the python package scikit-opt~\cite{skopt}. The hyper-parameters of these algorithms are chosen to be the values recommended by scikit-opt. Population size of 50 are used for DE and PSO. The inertia weight $w$, cognitive parameter $c1$, social parameter $c2$ are set to be 0.8, 0.5, 0.5. The $F$ of DE is chosen to be 0.5. In SA, min temperature, max temperature, long of chain, cooldown time are set to be 1e-7, 100, 300, 150. The experimental results of CV1.0 and dYYPO are extracted from their respective papers. One-tail two-sample t-test is applied to assess the relative performance of our proposed methods against CV1.0 and dYYPO. The outcome of the comparison is provided in Table~\ref{table:main}.

\begin{table*}[htbp]
\scriptsize
\centering
\caption{Brief comparison between algorithms}
\begin{tabular}{|c|c|c|c|c|c|c|} 
\hline
Properties & DE & PSO  & SA & CV1.0  & dYYPO  & YI   \\ 
\hline
Inspiration & \begin{tabular}{@{}c@{}}Genetic\\ evolution\end{tabular}  
& \begin{tabular}{@{}c@{}}Swarm of\\ birds\end{tabular}   
& \begin{tabular}{@{}c@{}}Metallurgy\\ in physics\end{tabular}    
& \begin{tabular}{@{}c@{}}Breeding of\\ Cuckoo\end{tabular}
& \begin{tabular}{@{}c@{}}Yin-yang\\ concept\end{tabular}  
& \begin{tabular}{@{}c@{}} Yi Jing\\ concept\end{tabular} \\
Solution representation & Vector
& Bird   
& Physics state   
& Cuckoo
& Yin / yang & Taiji \\
Population based &  Yes & Yes & Yes &  Yes   &  No     & No               \\
No of initial points & pop. size & pop. size  & pop. size & pop. size & Two & One\\
User defined parameter & Three & Four & Four & Two(trivial) & Three & Three\\
function eval. in step & pop. size & pop. size & pop. size & pop. size & $4D$ & $2D$\\
\hline
\end{tabular}
\label{tab:compare}
\end{table*}

We use $+$ to signify that a baseline algorithm under
consideration is better than our proposed algorithm, $-$
to signify the opposite, and $=$ to signify either they are performing similarly without statistical significance or the comparison is irrelevant. From the last row of Table~\ref{table:main}, the total count of the win, tie, loss (w/t/l) of DE, PSO, SA, CV1.0 and dYYPO against YI are shown. It is observed that the proposed algorithm is better than DE, PSO, SA, CV1.0 and dYYPO the majority of the time. YI won CV1.0 for 23 out 29 functions, won dYYPO for 21 out of 29 functions, won DE for 19 out of 29 functions, won PSO for all functions, and won SA for 24 out of 29 functions. The only irrelevant comparison was found when comparing dYYPO with YI for F3. YI was found to be much better even by comparing their mean and standard deviation; however, the standard deviation of dYYPO was too large to make statistical relevance. If dYYPO had obtained a more stable and reliable result, YI would win dYYPO for 22 out of 29 functions.

In addition, YI is generally found to obtain a relatively smaller standard deviation across runs (Table~\ref{table:main}), which demonstrates the stability of YI in convergence. Despite the computational simplicity of YI, it achieved a good balance between exploration and exploitation in the benchmark.

\subsection{Time Complexity}
The time complexity is calculated using the testing scheme provided in the CEC2017~\cite{Awad2016}. $T_0$ is counted by 1,000,000 times of basic operations, including addition, multiplication, exponentiation. $T_1$ is counted by running 200,000 times the optimization function F18. $T_2$ is counted by first running five trials of the Yi algorithm with the maximum number of function evaluation as 200,000, then averaging between the trials. The time complexity of the algorithm is calculated by $(T_2- T_1)/T_0$. We summarize the time complexity in Table~\ref{time}.

\begin{table}[htbp]
\centering
\caption{Time complexity~($T_0=11.61$)}
\begin{tabular}{|c|c|c|c|} 
\hline
D  & $T_1$  & $T_2$  & $(T_2-T_1)/T_0$   \\ 
\hline
10 &   4.50     &   24.44     &       1.72            \\
30 &    6.53    &   27.32     &     1.79          \\
50 &    8.95    &   30.47     &     1.85              \\
\hline
\end{tabular}
\label{time}
\end{table}

\begin{table}[]
\centering
\caption{Statistical Results for 10D}
\begin{tabular}{|c|r|r|r|r|}
\hline
\textbf{}    & \multicolumn{1}{c|}{\textbf{Best}} & \multicolumn{1}{c|}{\textbf{Worst}} & \multicolumn{1}{c|}{\textbf{Mean}} & \multicolumn{1}{c|}{\textbf{Std}} \\ \hline
\textbf{F1}  & 7.65E-01                           & 1.07E+04                            & 2.62E+03                           & 2.60E+03                          \\ \hline
\textbf{F3}  & 1.14E-06                           & 4.97E-06                            & 2.81E-06                           & 8.87E-07                          \\ \hline
\textbf{F4}  & 7.90E-03                           & 3.01E-02                            & 1.83E-02                           & 5.05E-03                          \\ \hline
\textbf{F5}  & 3.98E+00                           & 3.08E+01                            & 1.15E+01                           & 5.36E+00                          \\ \hline
\textbf{F6}  & 5.20E-04                           & 2.46E-03                            & 8.21E-04                           & 2.60E-04                          \\ \hline
\textbf{F7}  & 3.82E+00                           & 3.20E+01                            & 2.13E+01                           & 5.51E+00                          \\ \hline
\textbf{F8}  & 1.99E+00                           & 2.29E+01                            & 9.91E+00                           & 4.66E+00                          \\ \hline
\textbf{F9}  & 2.81E-07                           & 8.82E-07                            & 5.84E-07                           & 1.15E-07                          \\ \hline
\textbf{F10} & 3.60E+00                           & 6.96E+02                            & 2.89E+02                           & 1.88E+02                          \\ \hline
\textbf{F11} & 1.17E+00                           & 2.03E+01                            & 7.88E+00                           & 4.82E+00                          \\ \hline
\textbf{F12} & 9.39E+02                           & 4.98E+04                            & 1.43E+04                           & 1.31E+04                          \\ \hline
\textbf{F13} & 4.58E+02                           & 1.73E+04                            & 5.52E+03                           & 4.88E+03                          \\ \hline
\textbf{F14} & 2.20E+01                           & 9.46E+01                            & 4.40E+01                           & 1.64E+01                          \\ \hline
\textbf{F15} & 1.17E+01                           & 1.28E+02                            & 5.22E+01                           & 2.62E+01                          \\ \hline
\textbf{F16} & 1.06E+00                           & 1.20E+02                            & 5.20E+00                           & 1.63E+01                          \\ \hline
\textbf{F17} & 2.03E+00                           & 3.88E+01                            & 2.39E+01                           & 9.32E+00                          \\ \hline
\textbf{F18} & 1.26E+02                           & 2.15E+04                            & 5.39E+03                           & 5.75E+03                          \\ \hline
\textbf{F19} & 3.25E+00                           & 5.64E+01                            & 2.13E+01                           & 1.35E+01                          \\ \hline
\textbf{F20} & 1.00E+00                           & 4.60E+01                            & 2.04E+01                           & 9.75E+00                          \\ \hline
\textbf{F21} & 1.00E+02                           & 2.27E+02                            & 1.57E+02                           & 5.58E+01                          \\ \hline
\textbf{F22} & 1.00E+02                           & 1.03E+02                            & 1.02E+02                           & 6.78E-01                          \\ \hline
\textbf{F23} & 3.03E+02                           & 3.24E+02                            & 3.12E+02                           & 4.49E+00                          \\ \hline
\textbf{F24} & 1.22E-02                           & 3.53E+02                            & 2.90E+02                           & 9.01E+01                          \\ \hline
\textbf{F25} & 3.98E+02                           & 4.46E+02                            & 4.16E+02                           & 2.24E+01                          \\ \hline
\textbf{F26} & 1.46E-02                           & 3.00E+02                            & 2.65E+02                           & 5.88E+01                          \\ \hline
\textbf{F27} & 3.89E+02                           & 3.98E+02                            & 3.93E+02                           & 2.25E+00                          \\ \hline
\textbf{F28} & 3.00E+02                           & 6.12E+02                            & 3.06E+02                           & 4.32E+01                          \\ \hline
\textbf{F29} & 2.33E+02                           & 2.78E+02                            & 2.47E+02                           & 1.12E+01                          \\ \hline
\textbf{F30} & 1.03E+03                           & 1.24E+06                            & 1.05E+05                           & 3.11E+05                          \\ \hline
\end{tabular}
\label{tab:10D}
\end{table}

\begin{table}[]
\centering
\caption{Statistical Results for 30D}
\begin{tabular}{|c|r|r|r|r|}
\hline
\textbf{}    & \multicolumn{1}{c|}{\textbf{Best}} & \multicolumn{1}{c|}{\textbf{Worst}} & \multicolumn{1}{c|}{\textbf{Mean}} & \multicolumn{1}{c|}{\textbf{Std}} \\ \hline
\textbf{F1}  & 1.91E+02                           & 2.11E+04                            & 6.63E+03                           & 6.38E+03                          \\ \hline
\textbf{F3}  & 6.01E-04                           & 1.50E-03                            & 9.73E-04                           & 1.78E-04                          \\ \hline
\textbf{F4}  & 5.21E+00                           & 1.19E+02                            & 8.18E+01                           & 2.17E+01                          \\ \hline
\textbf{F5}  & 3.48E+01                           & 1.03E+02                            & 6.22E+01                           & 1.54E+01                          \\ \hline
\textbf{F6}  & 6.20E-03                           & 3.04E-01                            & 3.78E-02                           & 6.15E-02                          \\ \hline
\textbf{F7}  & 6.82E+01                           & 1.58E+02                            & 9.85E+01                           & 1.81E+01                          \\ \hline
\textbf{F8}  & 3.98E+01                           & 1.08E+02                            & 7.46E+01                           & 1.60E+01                          \\ \hline
\textbf{F9}  & 9.75E-05                           & 1.75E+01                            & 1.81E+00                           & 3.05E+00                          \\ \hline
\textbf{F10} & 1.35E+03                           & 3.13E+03                            & 2.34E+03                           & 4.41E+02                          \\ \hline
\textbf{F11} & 3.10E+01                           & 1.64E+02                            & 8.35E+01                           & 3.20E+01                          \\ \hline
\textbf{F12} & 2.82E+04                           & 1.26E+06                            & 3.45E+05                           & 2.89E+05                          \\ \hline
\textbf{F13} & 8.81E+03                           & 1.02E+05                            & 4.34E+04                           & 2.22E+04                          \\ \hline
\textbf{F14} & 1.75E+02                           & 1.15E+04                            & 1.88E+03                           & 2.20E+03                          \\ \hline
\textbf{F15} & 7.99E+03                           & 5.66E+04                            & 2.46E+04                           & 1.29E+04                          \\ \hline
\textbf{F16} & 5.65E+01                           & 1.01E+03                            & 4.83E+02                           & 2.11E+02                          \\ \hline
\textbf{F17} & 3.28E+01                           & 2.41E+02                            & 1.13E+02                           & 5.92E+01                          \\ \hline
\textbf{F18} & 1.15E+04                           & 2.18E+05                            & 7.80E+04                           & 4.19E+04                          \\ \hline
\textbf{F19} & 7.22E+02                           & 5.30E+04                            & 1.08E+04                           & 1.25E+04                          \\ \hline
\textbf{F20} & 3.38E+01                           & 3.24E+02                            & 1.38E+02                           & 7.63E+01                          \\ \hline
\textbf{F21} & 2.40E+02                           & 3.10E+02                            & 2.66E+02                           & 1.39E+01                          \\ \hline
\textbf{F22} & 1.00E+02                           & 1.05E+02                            & 1.01E+02                           & 1.39E+00                          \\ \hline
\textbf{F23} & 3.80E+02                           & 4.47E+02                            & 4.12E+02                           & 1.87E+01                          \\ \hline
\textbf{F24} & 4.58E+02                           & 5.59E+02                            & 4.95E+02                           & 1.93E+01                          \\ \hline
\textbf{F25} & 3.83E+02                           & 3.87E+02                            & 3.87E+02                           & 9.24E-01                          \\ \hline
\textbf{F26} & 2.00E+02                           & 2.18E+03                            & 1.56E+03                           & 4.34E+02                          \\ \hline
\textbf{F27} & 4.74E+02                           & 5.35E+02                            & 5.08E+02                           & 1.32E+01                          \\ \hline
\textbf{F28} & 3.00E+02                           & 4.89E+02                            & 3.36E+02                           & 5.73E+01                          \\ \hline
\textbf{F29} & 4.53E+02                           & 7.64E+02                            & 5.43E+02                           & 6.56E+01                          \\ \hline
\textbf{F30} & 9.13E+03                           & 4.10E+04                            & 2.39E+04                           & 7.94E+03                          \\ \hline
\end{tabular}
\label{table:30D}
\end{table}

\begin{table}[]
\centering
\caption{Statistical Results for 50D}
\begin{tabular}{|c|r|r|r|r|}
\hline
\textbf{}    & \multicolumn{1}{c|}{\textbf{Best}} & \multicolumn{1}{c|}{\textbf{Worst}} & \multicolumn{1}{c|}{\textbf{Mean}} & \multicolumn{1}{c|}{\textbf{Std}} \\ \hline
\textbf{F1}  & 2.17E+03                           & 3.37E+04                            & 8.94E+03                           & 7.07E+03                          \\ \hline
\textbf{F3}  & 7.90E-03                           & 1.69E-02                            & 1.23E-02                           & 1.78E-03                          \\ \hline
\textbf{F4}  & 2.85E+01                           & 2.24E+02                            & 1.26E+02                           & 4.68E+01                          \\ \hline
\textbf{F5}  & 7.66E+01                           & 2.41E+02                            & 1.47E+02                           & 3.03E+01                          \\ \hline
\textbf{F6}  & 2.74E-02                           & 1.21E+00                            & 2.06E-01                           & 2.29E-01                          \\ \hline
\textbf{F7}  & 1.28E+02                           & 3.04E+02                            & 2.04E+02                           & 3.82E+01                          \\ \hline
\textbf{F8}  & 9.55E+01                           & 2.06E+02                            & 1.40E+02                           & 2.36E+01                          \\ \hline
\textbf{F9}  & 1.91E+00                           & 6.05E+02                            & 8.46E+01                           & 1.22E+02                          \\ \hline
\textbf{F10} & 3.09E+03                           & 6.44E+03                            & 4.58E+03                           & 7.28E+02                          \\ \hline
\textbf{F11} & 8.22E+01                           & 3.04E+02                            & 1.68E+02                           & 4.69E+01                          \\ \hline
\textbf{F12} & 2.31E+05                           & 8.72E+06                            & 3.14E+06                           & 1.83E+06                          \\ \hline
\textbf{F13} & 2.04E+04                           & 1.38E+05                            & 6.03E+04                           & 2.48E+04                          \\ \hline
\textbf{F14} & 4.80E+02                           & 4.61E+04                            & 1.49E+04                           & 1.32E+04                          \\ \hline
\textbf{F15} & 5.34E+03                           & 6.01E+04                            & 2.68E+04                           & 1.25E+04                          \\ \hline
\textbf{F16} & 5.10E+02                           & 1.70E+03                            & 1.03E+03                           & 2.87E+02                          \\ \hline
\textbf{F17} & 4.29E+02                           & 1.22E+03                            & 7.36E+02                           & 2.04E+02                          \\ \hline
\textbf{F18} & 3.23E+04                           & 3.41E+05                            & 1.32E+05                           & 7.04E+04                          \\ \hline
\textbf{F19} & 1.13E+03                           & 4.18E+04                            & 1.63E+04                           & 1.32E+04                          \\ \hline
\textbf{F20} & 1.58E+02                           & 8.96E+02                            & 4.82E+02                           & 1.93E+02                          \\ \hline
\textbf{F21} & 2.91E+02                           & 4.19E+02                            & 3.42E+02                           & 2.92E+01                          \\ \hline
\textbf{F22} & 1.00E+02                           & 6.64E+03                            & 2.41E+03                           & 2.59E+03                          \\ \hline
\textbf{F23} & 4.82E+02                           & 6.27E+02                            & 5.62E+02                           & 2.86E+01                          \\ \hline
\textbf{F24} & 5.78E+02                           & 7.10E+02                            & 6.50E+02                           & 3.12E+01                          \\ \hline
\textbf{F25} & 4.60E+02                           & 5.78E+02                            & 5.09E+02                           & 3.44E+01                          \\ \hline
\textbf{F26} & 2.08E+03                           & 3.77E+03                            & 2.72E+03                           & 3.46E+02                          \\ \hline
\textbf{F27} & 5.08E+02                           & 7.77E+02                            & 5.65E+02                           & 4.09E+01                          \\ \hline
\textbf{F28} & 4.59E+02                           & 5.10E+02                            & 4.67E+02                           & 1.63E+01                          \\ \hline
\textbf{F29} & 3.96E+02                           & 1.18E+03                            & 7.40E+02                           & 1.97E+02                          \\ \hline
\textbf{F30} & 8.96E+05                           & 1.94E+06                            & 1.41E+06                           & 2.57E+05                          \\ \hline
\end{tabular}
\label{table:50D}
\end{table}

\begin{table}[]
\scriptsize
\centering
\caption{Statistical results of the proposed algorithm in comparison with
others}
\begin{tabular}{|l @{\hspace{1.5\tabcolsep}} l|l @{\hspace{1.5\tabcolsep}} l|l @{\hspace{1.5\tabcolsep}} l|l @{\hspace{1.5\tabcolsep}} l|l @{\hspace{1.5\tabcolsep}} l|l @{\hspace{1.5\tabcolsep}} l|l|}
\hline
                 &   &   & \textbf{CV1.0}    & \textbf{} & \textbf{dYYPO}    & \textbf{} & \textbf{DE}       & \textbf{} & \textbf{PSO}      & \textbf{} & \textbf{SA}       & \textbf{YI} \\ \hline
\textbf{F1}      & mean & - & 1.00E+10          & +         & 6.60E+03          & +         & \textbf{1.92E+03}          & -         & 1.21E+11          & -         & 2.79E+08          & 8.94E+03    \\
\textbf{}        & std  &   & 0.00E+00          &           & 7.10E+03          &           & \textbf{1.74E+03}          &           & 3.74E+10          &           & 5.17E+07          & 7.07E+03    \\
\textbf{F3}      & mean & - & 1.95E+04          & =         & 4.70E+01          & -         & 8.60E+04          & -         & 2.63E+05          & -         & 6.70E+04          & \textbf{1.23E-02}    \\
\textbf{}        & std  &   & 6.27E+03          &           & 2.30E+02          &           & 9.77E+03          &           & 8.54E+04          &           & 1.60E+04          & \textbf{1.78E-03}    \\
\textbf{F4}      & mean & = & 1.16E+02          & =         & 1.40E+02          & +         & \textbf{5.19E+01 }         & -         & 2.50E+04          & +         & 1.08E+02          & 1.26E+02    \\
\textbf{}        & std  &   & 6.27E+03          &           & 5.00E+01          &           & \textbf{3.70E+01}          &           & 1.24E+04          &           & 5.25E+01          & 4.68E+01    \\
\textbf{F5}      & mean & - & 3.41E+02          & -         & 1.90E+02          & -         & 3.16E+02          & -         & 7.20E+02          & -         & 3.91E+02          & \textbf{1.47E+02}    \\
\textbf{}        & std  &   & 8.02E+01          &           & 3.90E+01          &           & 1.38E+01          &           & 1.03E+02          &           & 4.33E+01          & \textbf{3.03E+01}    \\
\textbf{F6}      & mean & - & 4.85E+01          & -         & 3.80E+00          & +         & \textbf{1.14E-13}          & -         & 8.28E+01          & -         & 5.57E+01          & 2.06E-01    \\
\textbf{}        & std  &   & 4.85E+01          &           & 2.00E+00          &           & \textbf{0.00E+00}          &           & 1.09E+01          &           & 1.22E+01          & 2.29E-01    \\
\textbf{F7}      & mean & - & 2.74E+02          & -         & 2.60E+02          & -         & 3.70E+02          & -         & 3.70E+03          & -         & 5.40E+02          & \textbf{2.04E+02}    \\
\textbf{}        & std  &   & 7.29E+01          &           & 4.30E+01          &           & 1.52E+01          &           & 5.00E+02          &           & 6.64E+01          & \textbf{3.82E+01}    \\
\textbf{F8}      & mean & - & 3.29E+02          & -         & 1.90E+02          & -         & 3.16E+02          & -         & 7.42E+02          & -         & 3.95E+02          & \textbf{1.40E+02}    \\
\textbf{}        & std  &   & 7.29E+01          &           & 4.70E+01          &           & 1.54E+01          &           & 1.13E+02          &           & 6.16E+01          & \textbf{2.36E+01}    \\
\textbf{F9}      & mean & - & 1.00E+04          & -         & 3.50E+03          & +         & \textbf{6.02E-14}          & -         & 2.26E+04          & -         & 2.62E+04          & 8.46E+01    \\
\textbf{}        & std  &   & 2.90E+03          &           & 1.90E+03          &           & \textbf{5.67E-14}          &           & 4.90E+03          &           & 5.84E+03          & 1.22E+02    \\
\textbf{F10}     & mean & - & 7.10E+03          & =         & 4.80E+03          & -         & 1.16E+04          & -         & 8.86E+03          & -         & 8.21E+03          & \textbf{4.58E+03}    \\
\textbf{}        & std  &   & 5.34E+02          &           & 6.40E+02          &           & 3.76E+02          &           & 1.28E+03          &           & 8.15E+02          & \textbf{7.28E+02}    \\
\textbf{F11}     & mean & = & 1.66E+02          & -         & 1.90E+02          & =         & \textbf{1.57E+02}          & -         & 7.89E+03          & -         & 3.91E+02          & 1.68E+02    \\
\textbf{}        & std  &   & 3.38E+01          &           & 5.20E+01          &           & \textbf{1.22E+01}          &           & 7.29E+03          &           & 8.36E+01          & 4.69E+01    \\
\textbf{F12}     & mean & - & 1.00E+10          & -         & 7.80E+06          & -         & 8.50E+06          & -         & 3.30E+10          & -         & 5.72E+07          & \textbf{3.14E+06}    \\
\textbf{}        & std  &   & 0.00E+00          &           & 5.10E+06          &           & 3.50E+06          &           & 1.58E+10          &           & 3.43E+07          & \textbf{1.83E+06}    \\
\textbf{F13}     & mean & - & 1.00E+10          & +         & \textbf{7.60E+03}          & +         & 7.97E+03          & -         & 1.36E+10          & -         & 4.23E+05          & 6.03E+04    \\
\textbf{}        & std  &   & 0.00E+00          &           & \textbf{7.40E+03}          &           & 6.22E+03          &           & 8.60E+09          &           & 2.71E+05          & 2.48E+04    \\
\textbf{F14}     & mean & + & 2.05E+02          & -         & 2.90E+04          & -         & 2.37E+05          & -         & 6.01E+06          & -         & 4.89E+05          & \textbf{1.49E+04}    \\
\textbf{}        & std  &   & 2.13E+01          &           & 2.80E+04          &           & 1.38E+05          &           & 6.70E+06          &           & 3.74E+05          & \textbf{1.32E+04}    \\
\textbf{F15}     & mean & - & 1.37E+09          & +         & 8.20E+03          & +         & \textbf{3.08E+03}          & -         & 2.20E+09          & -         & 3.88E+04          & 2.68E+04    \\
\textbf{}        & std  &   & 3.47E+09          &           & 7.00E+03          &           & \textbf{1.35E+03}          &           & 3.77E+09          &           & 2.52E+04          & 1.25E+04    \\
\textbf{F16}     & mean & - & 1.53E+03          & -         & 1.30E+03          & -         & 1.80E+03          & -         & 3.66E+03          & -         & 1.67E+03          & \textbf{1.03E+03}    \\
\textbf{}        & std  &   & 2.74E+02          &           & 4.10E+02          &           & 1.78E+02          &           & 8.51E+02          &           & 4.52E+02          & \textbf{2.87E+02}    \\
\textbf{F17}     & mean & - & 1.25E+03          & -         & 8.90E+02          & -         & 8.47E+02          & -         & 1.11E+04          & -         & 1.43E+03          & \textbf{7.36E+02}\\
\textbf{}        & std  &   & 1.85E+02          &           & 2.70E+02          &           & 1.24E+02          &           & 2.33E+04          &           & 3.56E+02          & \textbf{2.04E+02}    \\
\textbf{F18}     & mean & + & \textbf{5.21E+02}          & -         & 1.80E+05          & -         & 2.00E+06          & -         & 5.15E+07          & -         & 3.11E+06          & 1.32E+05    \\
\textbf{}        & std  &   & \textbf{1.19E+02}          &           & 8.10E+04          &           & 8.27E+05          &           & 8.79E+07          &           & 2.47E+06          & 7.04E+04    \\
\textbf{F19}     & mean & + & \textbf{1.73E+02}          & +         & 9.60E+03          & +         & 7.55E+03          & -         & 1.14E+09          & -         & 2.08E+04          & 1.63E+04    \\
\textbf{}        & std  &   & \textbf{4.17E+02}          &           & 8.80E+03          &           & 3.57E+03          &           & 1.54E+09          &           & 1.44E+04          & 1.32E+04    \\
\textbf{F20}     & mean & - & 1.05E+03          & -         & 6.50E+02          & -         & 6.49E+02          & -         & 1.69E+03          & -         & 1.06E+03          & \textbf{4.82E+02}    \\
\textbf{}        & std  &   & 2.14E+02          &           & 2.80E+02          &           & 1.31E+02          &           & 3.16E+02          &           & 2.72E+02          & \textbf{1.93E+02}    \\
\textbf{F21}     & mean & - & 5.41E+02          & -         & 4.00E+02          & -         & 5.21E+02          & -         & 9.04E+02          & -         & 6.09E+02          & \textbf{3.42E+02}    \\
\textbf{}        & std  &   & 6.27E+01          &           & 4.00E+01          &           & 1.21E+01          &           & 1.11E+02          &           & 6.97E+01          & \textbf{2.92E+01}    \\
\textbf{F22}     & mean & - & 7.33E+03          & -         & 4.80E+03          & -         & 1.05E+04          & -         & 9.31E+03          & -         & 8.01E+03          & \textbf{2.41E+03}    \\
\textbf{}        & std  &   & 1.99E+03          &           & 2.00E+03          &           & 2.99E+03          &           & 1.25E+03          &           & 3.22E+03          & \textbf{2.59E+03}    \\
\textbf{F23}     & mean & - & 7.74E+02          & -         & 6.50E+02          & -         & 7.38E+02          & -         & 1.36E+03          & -         & 9.59E+02          & \textbf{5.62E+02}    \\
\textbf{}        & std  &   & 8.06E+01          &           & 5.70E+01          &           & 1.42E+01          &           & 1.77E+02          &           & 9.08E+01          & \textbf{2.86E+01}    \\
\textbf{F24}     & mean & - & 8.32E+02          & -         & 7.30E+02          & -         & 8.45E+02          & -         & 1.37E+03          & -         & 1.02E+03          & \textbf{6.50E+02}    \\
\textbf{}        & std  &   & 1.21E+01          &           & 7.40E+01          &           & 1.06E+01          &           & 1.68E+02          &           & 9.47E+01          & \textbf{3.12E+01}    \\
\textbf{F25}     & mean & - & 5.43E+02          & -         & 5.20E+02          & =         & 5.17E+02          & -         & 1.57E+04          & =         & \textbf{5.03E+02}          & 5.09E+02    \\
\textbf{}        & std  &   & 1.51E+01          &           & 3.00E+01          &           & 3.77E+01          &           & 6.70E+03          &           & \textbf{4.88E+01}          & 3.44E+01    \\
\textbf{F26}     & mean & = & 2.48E+03          & -         & 3.40E+03          & -         & 4.08E+03          & -         & 1.19E+04          & +         & \textbf{9.57E+02}          & 2.72E+03    \\
\textbf{}        & std  &   & 1.88E+03          &           & 7.80E+02          &           & 1.06E+02          &           & 2.25E+03          &           & \textbf{1.24E+03}          & 3.46E+02    \\
\textbf{F27}     & mean & - & 7.38E+02          & -         & 6.70E+02          & +         & 5.50E+02          & -         & 1.59E+03          & +         & \textbf{4.91E+02}          & 5.65E+02    \\
\textbf{}        & std  &   & 8.21E+01          &           & 7.30E+01          &           & 1.27E+01          &           & 3.48E+02          &           & \textbf{1.35E+01}          & 4.09E+01    \\
\textbf{F28}     & mean & - & 4.94E+02          & -         & 4.80E+02          & -         & 4.76E+02          & -         & 9.39E+03          & -         & 4.97E+02          & \textbf{4.67E+02}    \\
\textbf{}        & std  &   & 1.93E+01          &           & 2.50E+01          &           & 2.23E+01          &           & 1.80E+03          &           & 3.03E+01          & \textbf{1.63E+01}    \\
\textbf{F29}     & mean & - & 1.69E+03          & -         & 9.80E+02          & -         & 1.09E+03          & -         & 5.82E+03          & -         & 1.69E+03          & \textbf{7.40E+02}    \\
\textbf{}        & std  &   & 2.29E+02          &           & 3.10E+02          &           & 1.37E+02          &           & 3.01E+03          &           & 3.98E+02          & \textbf{1.97E+02}    \\
\textbf{F30}     & mean & - & 4.64E+06          & =         & 1.50E+06          & -         & 2.25E+06          & -         & 1.87E+09          & +         & \textbf{8.90E+05}          & 1.41E+06    \\
\textbf{}        & std  &   & 8.59E+06          &           & 3.20E+05          &           & 3.60E+05          &           & 1.70E+09          &           & \textbf{3.98E+05}          & 2.57E+05    \\ \hline
\multicolumn{2}{|c|}{\textbf{(w,t,l)}} &  & \textbf{(3,3,23)} & \textbf{} & \textbf{(4,4,21)} & \textbf{} & \textbf{(8,2,19)} & \textbf{} & \textbf{(0,0,29)} & \textbf{} & \textbf{(4,1,24)} &  \\ \hline         
\end{tabular}
\label{table:main}
\end{table}

\begin{table}[]
\centering
\caption{Parameter analysis\\(With the case $\alpha=3$,$I_{min}=6$ and $I_{max}=15$)}
\begin{tabular}{|c|c|c|c|c|c|}
\hline
\textbf{$I_{min}$} & \textbf{$I_{max}$} & \textbf{$\sigma$} & \textbf{Win} & \textbf{Tie} & \textbf{Loss} \\ \hline
6    & 15   & 1.5   & 0   & 1   & 28   \\
6    & 15   & 5     & 2   & 13  & 14   \\
1    & 10   & 3     & 2   & 21  & 6    \\
16   & 25   & 3     & 6   & 21  & 2    \\
6    & 25   & 3     & 0   & 21  & 8   \\ \hline
\end{tabular}
\label{parameter}
\end{table}

\section{Discussions}
In this section, we discuss the main experimental results of the proposed method, in terms of the convergence to a good solution and time complexity. Based on the results, we also discuss the concept of YI in view of its simplicity and competitive performance against DE, PSO, SA, dYYPO and CV1.0. Generally, YI performs consistently well in different types of functions. DE performs comparably well with YI in unimodal and multi-modal functions F1-F10, while SA performs comparably well with YI in composition function F21-F30.

\begin{itemize}
    \item Convergence and handling local minima: In the unimodal functions, YI performed very well in F3 Zakharov function, converging to high precision near the global minimum in all the dimensions. Its error in means is three orders of magnitude smaller than dYYPO in 50D. The result indicates the exploitation task is done much better in YI than in dYYPO. However, YI does not outperform dYYPO in F1 Bent Cigar function, where the narrow ridge of local minimum exists; this might be due to the multi-strategy of splitting in dYYPO, enabling it performs better search along the ridge. In the multi-modal functions F4 to F10, YI outperformed both dYYPO and CV1.0 in most cases. It demonstrated the competitiveness of YI in escaping different types of local minimums. Especially in F9 $\mathrm{L\acute{e}vy}$ Function, where a huge number of local minimum presents, Yi can achieve two orders of magnitude better in error. YI is highly competitive with DE, except for the function where the ridge or the large area of plateau exist, like F1 Bent Cigar function, F4 Rosenbrock function, F6 Schaffer’s Function and F9 Levy function. Also, YI outperforms both PSO and SA in most of the functions.

    \item Hybrid function: The hybrid functions are complicated functions involving different basic functions for different subcomponents of the variable. The efficacy of search in these functions highly depends on the searching strategy. YI and dYYPO share many similarities in the strategy design, while CV1.0 is in a completely different realm. We expect the performance of YI and dYYPO are mostly consistent. Although YI still performs better in many cases comparing to dYYPO and CV1.0, it does not significantly outperform dYYPO in any of the hybrid functions. In F13, F15, F19, dYYPO and DE is better than YI significantly. In these three hybrid functions, the F1 Bent Cigar function is involved; this could result in the weak performance of YI.
    
    \item Composition functions: YI performs consistently better in the composition function than DE, PSO, dYYPO and CV1.0, in which sub-functions properties are merged. SA is competitive with YI in handling composition function, as SA optimizes to a global minimum by decreasing temperature repeatedly, which is advantageous in handling composite function in which the different functions are at different energy scales. Even without sharing algorithmic features with SA, YI performs decently in handling composition functions. This proves the YI algorithm can handle complicated function landscape very well to search the global minimum. 
    
    \item Dimensionality: When we compare the performance of YI with dYYPO~\cite{Maharana2017} in different dimensions, we find YI performs better in the higher dimensions. In 10D, YI is only slightly better than dYYPO in terms of smaller mean errors among all 29 functions. The advantage becomes more and more obvious when coming to 50D. This could be due to the Yi algorithm the complicated landscape in high dimensions. 
    
    \item Time complexity: As shown in Table~\ref{time}, similar to dYYPO, the time complexity of YI does not increase significantly with the dimensionality. This is beneficial when we perform searches in high dimensions. Especially the real-time problem today is complicated and often requires to search in high dimensional function space.
    
    \item Hyper parameters: We have tested different choices of hyper-parameters $\sigma$, $I_{min}$ and $I_{max}$. As listed in the Table \ref{parameter}, the decay rate $\sigma$ has a great influence on the optimization result. In the searching process, either decaying too fast~($\sigma=5$) or too slow ($\sigma=1.5$) deteriorates the performance of YI. We also find that the performance of YI is not very sensitive to the change in $I_{max}$ and $I_{min}$, which control the number of intervals and the archive time of each interval in the run.
    
\end{itemize}

\section{Conclusions}
The paper proposed a novel non-population-based algorithm for meta-heuristics optimization by drawing inspiration from the Yi Jing. Instead of using the Yin-Yang pair in YYPO, the concept of YI is incorporated into $\mathrm{L\acute{e}vy}$ flight for enhanced exploration and exploitation while maintaining a trade-off between both of them. The proposed method also inherits the concept of simplicity from the Yi Jing, which provides the advantage of low time complexity. This allows YI to be used in the high-dimensional optimization problems. 

According to the experimental results, our proposed methods demonstrated superior performance against DE,PSO, SA, CV1.0 and dYYPO on CEC 2017 benchmark that contains numerous types of challenging functions. The competitive results suggested that the proposed method achieved a good balance between exploration and exploitation. Comparing to the selected algorithms, YI is better at escaping local minima and handling complicated function landscapes in most cases. In the ridge type of function landscape like Bent Cigar function, YI performs relatively weak. This could be improved if we include other types of searching strategies, like the directional $\mathrm{L\acute{e}vy}$ flight as in dYYPO. Moreover, the Yi-point does not limit to the use $\mathrm{L\acute{e}vy}$ flight updating strategy. We can also adapt other strategies, like the Cauchy function, as long as being able to balance between the exploration and the exploitation task. 

While this work starts a new chapter for the Yi Jing inspired optimizer, the work can be extended in numerous ways. The extension of YI from single-objective optimization to multi-objective or many objective optimizations deserves research attention~\cite{tanabe2019review}. The concept of improved control over the exploration and exploitation in YI using the dynamical achieving and the YI point can fuse with other algorithms, including the population base algorithms~(e.g. PSO, GA) to generate an even more competitive optimizer.



 \bibliographystyle{elsarticle-num} 
 \bibliography{cas-refs}

\begin{thebibliography}{10}
\expandafter\ifx\csname url\endcsname\relax
  \def\url#1{\texttt{#1}}\fi
\expandafter\ifx\csname urlprefix\endcsname\relax\def\urlprefix{URL }\fi
\expandafter\ifx\csname href\endcsname\relax
  \def\href#1#2{#2} \def\path#1{#1}\fi

\bibitem{mirjalili2020genetic}
S.~Mirjalili, J.~S. Dong, A.~S. Sadiq, H.~Faris, Genetic algorithm: Theory,
  literature review, and application in image reconstruction, Nature-inspired
  optimizers (2020) 69--85.

\bibitem{Yang2009}
X.~S. Yang, S.~Deb, {Cuckoo search via L{\'{e}}vy flights}, in: 2009 World
  Congress on Nature and Biologically Inspired Computing - Proceedings, 2009,
  pp. 210--214.
\newblock \href {http://arxiv.org/abs/1003.1594} {\path{arXiv:1003.1594}}.

\bibitem{mirjalili2014grey}
S.~Mirjalili, S.~M. Mirjalili, A.~Lewis, Grey wolf optimizer, Advances in
  engineering software 69 (2014) 46--61.

\bibitem{brindle1980genetic}
A.~Brindle, Genetic algorithms for function optimization (1980).

\bibitem{al2019survey}
H.~Al-Sahaf, Y.~Bi, Q.~Chen, A.~Lensen, Y.~Mei, Y.~Sun, B.~Tran, B.~Xue,
  M.~Zhang, A survey on evolutionary machine learning, Journal of the Royal
  Society of New Zealand 49~(2) (2019) 205--228.

\bibitem{sinha2017review}
A.~Sinha, P.~Malo, K.~Deb, A review on bilevel optimization: from classical to
  evolutionary approaches and applications, IEEE Transactions on Evolutionary
  Computation 22~(2) (2017) 276--295.

\bibitem{fernandez2019evolutionary}
A.~Fernandez, F.~Herrera, O.~Cordon, M.~J. del Jesus, F.~Marcelloni,
  Evolutionary fuzzy systems for explainable artificial intelligence: Why,
  when, what for, and where to?, IEEE Computational Intelligence Magazine
  14~(1) (2019) 69--81.

\bibitem{Punnathanam2016}
V.~Punnathanam, P.~Kotecha, Reduced yin-yang-pair optimization and its
  performance on the cec 2016 expensive case, 2016 IEEE Congress on
  Evolutionary Computation, CEC 2016 (2016) 2996--3002.

\bibitem{Punnathanam2016a}
V.~Punnathanam, P.~Kotecha, Yin-yang-pair optimization: A novel lightweight
  optimization algorithm, Engineering Applications of Artificial Intelligence
  54 (2016) 62--79.

\bibitem{Maharana2017}
D.~Maharana, R.~Kommadath, P.~Kotecha, Dynamic yin-yang pair optimization and
  its performance on single objective real parameter problems of cec 2017, 2017
  IEEE Congress on Evolutionary Computation, CEC 2017 - Proceedings (2017)
  2390--2396.

\bibitem{Punnathanam2019}
V.~Punnathanam, P.~Kotecha, Optimization of multi-objective dynamic
  optimization problems with front-based yin-yang-pair optimization, in: Smart
  Innovations in Communication and Computational Sciences, Springer, 2019, pp.
  377--386.

\bibitem{Punnathanam2019a}
V.~Punnathanam, P.~Kotecha, Front-based yin-yang-pair optimization and its
  performance on cec2009 benchmark problems, in: Smart Innovations in
  Communication and Computational Sciences, Springer, 2019, pp. 387--397.

\bibitem{Punnathanam2017}
V.~Punnathanam, P.~Kotecha, {Multi-objective optimization of Stirling engine
  systems using Front-based Yin-Yang-Pair Optimization}, Energy Conversion and
  Management 133 (2017) 332--348.

\bibitem{Heidari2017}
A.~A. Heidari, O.~Kazemizade, F.~Hakimpour, {A new hybrid
  yin-yang-pair-particle swarm optimization algorithm for uncapacitated
  warehouse location problems}, Int. Arch. Photogramm. Remote Sens. Spat. Inf.
  Sci. - ISPRS Arch. 42~(4W4) (2017) 373--379.

\bibitem{Yang2018}
B.~Yang, T.~Yu, H.~Shu, D.~Zhu, F.~Zeng, Y.~Sang, L.~Jiang, {Perturbation
  observer based fractional-order PID control of photovoltaics inverters for
  solar energy harvesting via Yin-Yang-Pair optimization}, Energy Convers.
  Manag. 171~(March) (2018) 170--187.

\bibitem{Song2020}
D.~Song, J.~Liu, J.~Yang, M.~Su, Y.~Wang, X.~Yang, L.~Huang, Y.~H. Joo,
  {Optimal design of wind turbines on high-altitude sites based on improved
  Yin-Yang pair optimization}, Energy 193 (2020) 116794.

\bibitem{kennedy1995particle}
J.~Kennedy, R.~Eberhart, {Particle swarm optimization}, in: Proc.
  ICNN'95-international Conf. neural networks, Vol.~4, IEEE, 1995, pp.
  1942--1948.

\bibitem{shi1998modified}
Y.~Shi, R.~Eberhart, {A modified particle swarm optimizer}, in: 1998 IEEE Int.
  Conf. Evol. Comput. proceedings. IEEE world Congr. Comput. Intell. (Cat. No.
  98TH8360), IEEE, 1998, pp. 69--73.

\bibitem{wu2017problem}
G.~Wu, R.~Mallipeddi, P.~N. Suganthan, Problem definitions and evaluation
  criteria for the cec 2017 competition on constrained real-parameter
  optimization, Technical Report (2017).

\bibitem{yao1999evolutionary}
X.~Yao, Y.~Liu, G.~Lin, Evolutionary programming made faster, IEEE Transactions
  on Evolutionary computation 3~(2) (1999) 82--102.

\bibitem{brown2007levy}
C.~T. Brown, L.~S. Liebovitch, R.~Glendon, L{\'e}vy flights in dobe
  ju/’hoansi foraging patterns, Human Ecology 35~(1) (2007) 129--138.

\bibitem{pavlyukevich2007levy}
I.~Pavlyukevich, L{\'e}vy flights, non-local search and simulated annealing,
  Journal of Computational Physics 226~(2) (2007) 1830--1844.

\bibitem{storn1997differential}
R.~Storn, K.~Price, Differential evolution--a simple and efficient heuristic
  for global optimization over continuous spaces, Journal of global
  optimization 11~(4) (1997) 341--359.

\bibitem{10.3389/fbuil.2020.00102}
M.~Georgioudakis, V.~Plevris, A comparative study of differential evolution
  variants in constrained structural optimization, Frontiers in Built
  Environment 6 (2020) 102.

\bibitem{kirkpatrick1983optimization}
S.~Kirkpatrick, C.~D. Gelatt, M.~P. Vecchi, Optimization by simulated
  annealing, Science 220~(4598) (1983) 671--680.

\bibitem{jensi2016enhanced}
R.~Jensi, G.~W. Jiji, An enhanced particle swarm optimization with levy flight
  for global optimization, Applied Soft Computing 43 (2016) 248--261.

\bibitem{chegini2018psoscalf}
S.~N. Chegini, A.~Bagheri, F.~Najafi, Psoscalf: A new hybrid pso based on sine
  cosine algorithm and levy flight for solving optimization problems, Applied
  Soft Computing 73 (2018) 697--726.

\bibitem{abdulwahab2019enhanced}
H.~A. Abdulwahab, A.~Noraziah, A.~A. Alsewari, S.~Q. Salih, An enhanced version
  of black hole algorithm via levy flight for optimization and data clustering
  problems, IEEE Access 7 (2019) 142085--142096.

\bibitem{amirsadri2018levy}
S.~Amirsadri, S.~J. Mousavirad, H.~Ebrahimpour-Komleh, A levy flight-based grey
  wolf optimizer combined with back-propagation algorithm for neural network
  training, Neural Computing and Applications 30~(12) (2018) 3707--3720.

\bibitem{salgotra2018new}
R.~Salgotra, U.~Singh, S.~Saha, New cuckoo search algorithms with enhanced
  exploration and exploitation properties, Expert Systems with Applications 95
  (2018) 384--420.

\bibitem{skopt}
Scikit-optimize, \url{https://scikit-optimize.github.io/} (2021).

\bibitem{Awad2016}
N.~H. Awad, M.~Z. Ali, P.~N. Suganthan, J.~J. Liang, B.~Y. Qu, {Problem
  Definitions and Evaluation Criteria for the CEC 2017 Special Session and
  Competition on Single Objective Bound Constrained Real-Parameter Numerical
  Optimization}, 2016.

\bibitem{tanabe2019review}
R.~Tanabe, H.~Ishibuchi, A review of evolutionary multimodal multiobjective
  optimization, IEEE Transactions on Evolutionary Computation 24~(1) (2019)
  193--200.

\end{thebibliography}





\end{document}